\documentclass[fleqn]{article}
\usepackage{spconf,amsmath,amssymb,graphicx}
\usepackage{float, booktabs}
\usepackage{multirow}

\usepackage{pifont}
\newcommand{\xmark}{\ding{55}}

\title{M3T: Multi-Modal Medical Transformer to bridge Clinical Context with Visual Insights for Retinal Image Medical Description Generation}
\name{Nagur Shareef Shaik$^{*}$, Teja Krishna Cherukuri$^{*}$, Dong Hye Ye$^{*}$
}
\address{$^*$Department of Computer Science, Georgia State University, Atlanta, GA, United States}

\begin{document}

\maketitle

\begin{abstract}
Automated retinal image medical description generation is crucial for streamlining medical diagnosis and treatment planning. Existing challenges include the reliance on learned retinal image representations, difficulties in handling multiple imaging modalities, and the lack of clinical context in visual representations. Addressing these issues, we propose the Multi-Modal Medical Transformer (M3T), a novel deep learning architecture that integrates visual representations with diagnostic keywords. Unlike previous studies focusing on specific aspects, our approach efficiently learns contextual information and semantics from both modalities, enabling the generation of precise and coherent medical descriptions for retinal images. Experimental studies on the DeepEyeNet dataset validate the success of M3T in meeting ophthalmologists' standards, demonstrating a substantial 13.5\% improvement in BLEU@4 over the best-performing baseline model.
\end{abstract}

\begin{keywords}
Medical Description Generation, Multi-Modal Learning, Medical Image Analysis, Transformer
\end{keywords}

\vspace{-3mm}
\section{Introduction} \label{sec:introduction}
\vspace{-2mm}

In recent years, the global prevalence of visual impairments, such as Diabetic Retinopathy, Diabetic Macular Edema, and Age-related Macular Edema, has significantly increased. These conditions, often causing irreversible blindness, stem from complications like sub-retinal neo-vascularization and retinal detachment \cite{huang2021deepopht}. Predictions by the World Health Organization estimate that over 500 million people will be affected by these retinal diseases by 2040 \cite{zheng2012worldwide}. Traditional diagnosis relies on resource-intensive processes, expensive equipment, and the expertise of ophthalmologists, resulting in challenges for timely and accurate medical reporting. Automated generation of medical descriptions from retinal images holds potential for advancing early diagnosis \cite{shaik2024gated}. Aligned with Image Captioning, this task integrates Computer Vision and Natural Language Processing, employing Convolutional Neural Networks (CNNs) for image representations fed into Recurrent Neural Networks (RNNs) for textual description generation \cite{xu2015show, kamal2020textmage}. Despite breakthroughs in Natural Image Captioning using advanced models like Transformers, accurately captioning complex retinal images remains challenging due to the limited annotated medical data \cite{beddiar2022automatic}. The scarcity of labeled data poses challenges in both image classification and caption generation tasks in medical image analysis. Researchers address this by employing Transfer Learning, leveraging models pre-trained on ImageNet for medical image tasks \cite{shaik2021lesion, shaik2022hinge}. Pre-training on natural images and fine-tuning on medical datasets enhances feature learning, especially in medical image classification \cite{shaik2022transfer}. Semi-supervised and self-supervised learning in medical representation explores unlabeled data, benefiting subsequent tasks \cite{misra2020self, li2023dynamic}. Despite these advances, generating clinically relevant captions for retinal images remains challenging due to the complexity of interpreting retinal pathology \cite{huang2021deep}.

\begin{figure*}[!t]
    \centering
    \centerline{\includegraphics[width=.88\textwidth]{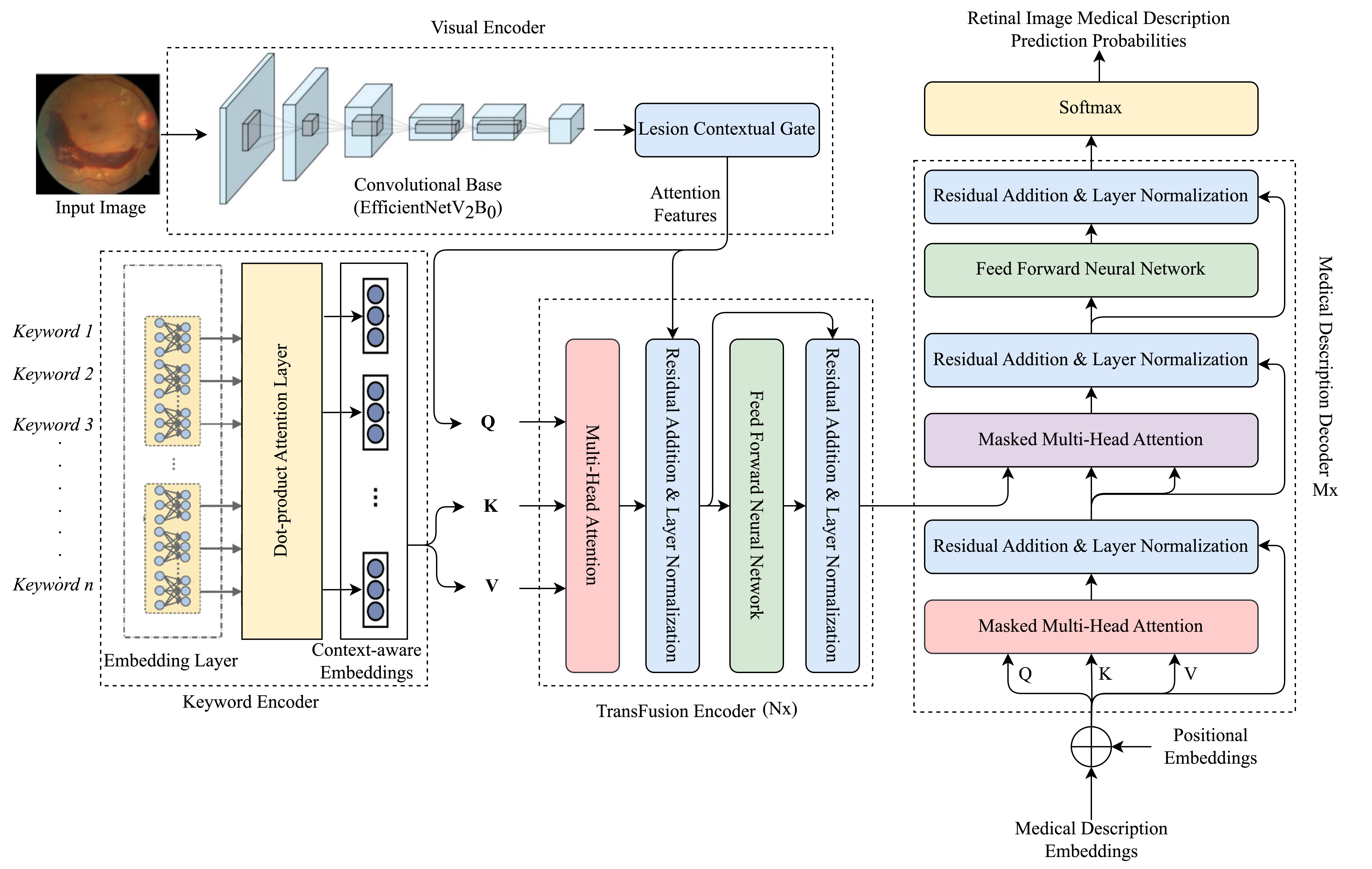}}
    \vspace{-5mm}
    \caption{Architecture of proposed Multi-Modal Medical Transformer (M3T); \textbf{Visual Encoder} -- learns attention-based representations from retinal images; \textbf{Keyword Encoder} -- learns clinical-context embeddings from diagnostic keywords; \textbf{TransFusion Encoder} -- integrates visual attention features and clinical-context embeddings, leveraging both visual and semantic information; \textbf{Medical Description Generation Decoder} -- generates coherent and meaningful medical descriptions by attending to relevant visual and semantic cues, ensuring contextually appropriate and diagnostically relevant outputs;}
    \vspace{-5mm}
    \label{fig:architecture}
\end{figure*} 

Natural Image Caption Generation has improved with multi-modal inputs \cite{herdade2019image}. For medical descriptions, integrating diagnostic keywords aids in capturing semantics, but efficiently learning visual context and clinical semantics presents challenges. The CLIP model, which aligns images and text in a shared embedding space, also offers significant potential for improving the integration of visual and textual information \cite{radford2021learning}. A deep learning framework utilizing both images \& keywords enhances performance in retinal disease diagnosis \cite{huang2021deepopht, huang2021deep}. Similarly, contextualized keyword embeddings derived from techniques like GloVe and GPT2 were integrated with retinal image features for caption generation, with GPT2 embeddings proving superior \cite{huang2021contextualized}. Despite these efforts, reported quantitative metrics were less competitive due to the choice of image features and the lack of domain-specific training. Notably, incorporating Non-local attention-based retinal image features with diagnostic keywords led to more semantically meaningful medical image captions, indicating improved performance \cite{huang2022non}. Further enhancement was achieved by training a Transformer model on visual features and keyword embeddings \cite{wu2023expert}.

Most of the aforementioned studies have concentrated on specific aspects, such as the type of diagnostic keyword embeddings or the attention mechanisms for visual features. However, there exists a significant gap in efficiently learning contextual information and semantics from both modalities and integrating them effectively. This research endeavors to bridge this gap by proposing a novel deep learning architecture called the Multi-Modal Medical Transformer. Our primary goal is to learn visual context from retinal images, understand clinical semantics from diagnostic keywords, and establish an effective integration mechanism through attention. By amalgamating these crucial components, our aim is to produce precise and coherent medical descriptions for retinal images.

\vspace{-5mm}
\section{Methodology} \label{sec:methods}
\vspace{-2mm}

The goal of this research is to induce a novel deep learning framework, Multi-Modal Medical Transformer that attentively integrates clinical context from diagnostic keywords with visual insights obtained from multi-modal retinal scan images for comprehensive medical description generation. The framework comprises multiple modules that handle the processing of multi-modal input and contribute to the generation of semantically coherent medical descriptions. Figure \ref{fig:architecture} illustrates the complete architecture of proposed model. Technical specifications for each module are elaborated in subsequent subsections.

\vspace{-3mm}
\subsection{Visual Encoder}
\vspace{-2mm}

The Visual Encoder is designed for converting the pre-processed retinal images into visual features through a special attention mechanism. This module is structured with a convolutional base and Lesion contextual gate operating collaboratively to extract lesion contextual visual representations. 

\vspace{-3mm}
\subsubsection{Convolutional Base}
\vspace{-2mm}

The Convolutional base serves to extract initial visual representations capturing essential patterns in images. Convolutional Neural Networks (CNNs) excel in learning spatial representations across various image modalities, but training them from scratch requires substantial labeled data, often limited in medical image tasks like retinal scans. To overcome this limitation, we utilize pre-trained CNN models, initially trained on ImageNet, benefiting from generalizable visual features applicable to retinal images. Specifically, we choose EfficientNet$V_2B_0$ as the Convolutional base, which integrates Squeeze and Excitation (SE) blocks for attentive channel-wise information consideration, to extract spatial representations from multi-modal retinal images.
Passing a retinal scan image $X_R$ with dimensions $(356 \times 356 \times 3)$ through EfficientNet $V_2B_0$ yields visual features $V$ of dimensions $(12 \times 12 \times 1280)$ from the final convolutional block.

\vspace{-3mm}
\subsubsection{Lesion Contextual Gate}
\vspace{-2mm}

The retinal representations obtained from the pre-trained EfficientNet $V_2B_0$ exhibit global information, where equal importance is assigned to each feature map. Despite the success of attention mechanisms like Squeeze and Excitation blocks, our model still faces challenges in capturing lesion-specific contextual information crucial for accurate diagnosis. To address this limitation, we introduce a Lesion Contextual Gate Attention block that enhances retinal image representations with lesion context-rich information by using single attention block that considers global and local context information.

Initially, lesion context information is computed through global attention pooling which involves a point-wise convolution capturing essential features, and a softmax activation normalizes spatial elements generating an attention map. This attention map selectively focuses on relevant features in the initial spatial representations, providing lesion context information. Next, channel-wise dependencies through specialized point-wise convolution layers which process previously computed context information, introducing non-linearity and discerning intricate dependencies among different channels. The resulting features are aggregated with original spatial features, enriching contextual understanding within the retinal features. Finally, the gating function enhances the model's focus on lesion-specific regions where coefficients are computed based on global and local context information, selectively amplifying or suppressing features in retinal representations. The integration of weighted representations with normalization, and gating ensures effective highlighting of lesion-specific details, capturing both global and localized context features within the retinal images. Below are the mathematical computations involved in Lesion Contextual Gate module. 
\begin{equation}
    F_{gap} = \sum\limits_{j=1}^d \frac{e^{W_c f_{r,j}}}{\sum\limits_{m=1}^d e^{W_c f_{r,m}}} \qquad \forall f_{r,j} \in F_R,
    \label{eq:gap}
\end{equation}
where $F_{gap}$ represents the global attention pooling of visual features $F_R$, $W_c$ represents the the parameters of point-wise convolution. The numerator computes the exponential of the weighted visual features, and the denominator is the sum of exponentiated weighted visual features by using a softmax activation, resulting in a normalized global context feature.
\begin{equation}
    F_{c} = F_R \oplus \sum\limits_{i=1}^d W_{2}\left(\text{LN}\left(\Gamma\left(\sum\limits_{j=1}^{k} W_{1}f_{g_j}\right)\right)\right)_i
    \forall f_{g_j} \in F_{gap},
    \label{eq:cc}
\end{equation}
where $F_c $ represents the channel attentive feature incorporating global context, $W_1, W_2$ represents the parameters of point-wise convolutions, $\Gamma$ is ReLU non-linearity and LN indicates Layer Normalization. The operation involves weighted summation and layer normalization of the global context features $F_{gap}$. This enriched features are then element-wise added to the original visual features $F_R$.
\begin{equation}
    F_{att} = \sigma(W_\psi \cdot \Gamma(W_x F_{R} + W_g F_c + b_{xg}) + b_{\psi}),
    \label{eq:gate}
\end{equation}
where $F_{att}$ represents the final attention-weighted visual features. This is computed by applying the gating mechanism using parameters  $W_x, W_g, b_{xg}, W_\psi, b_{\psi}$ . This gating is modulated by global context features from $F_c$ and original visual features $F_R$.

\vspace{-3mm}
\subsection{Keywords Encoder}
\vspace{-2mm}

The Keywords Encoder functions as a language modeling encoder, transforming processed diagnostic keywords into context-aware embeddings. Within the encoder module, the Embedding layer takes the processed diagnostic keyword sequence as input, converting it into numerical values known as embeddings (Word2Vec). These embeddings, denoted as $E$, encapsulate individual keyword embeddings $ke_1, ke_2, ..., ke_n$, effectively capturing the semantic relationships embedded in the input keywords.

However, the initially generated embeddings are context-free, lacking explicit consideration of relationships between keywords. To address this limitation, the embeddings undergo processing by an Attention layer. This layer leverages a dot product operation to compute alignment scores between source $(ke_s)$ and target $(ke_t)$ embeddings, denoted as $\bar{ke}$ \cite{luong2015effective}. These alignment scores reflect the importance or relevance of each keyword relative to others in the sequence. Subsequently, a Softmax activation function is applied to normalize the alignment scores, yielding attention weights $ke_{att}$. These weights furnish a context-aware representation of diagnostic keywords, assigning higher weights to more pertinent keywords and lower weights to less relevant ones. This mechanism enables the model to focus on the most informative keyword embeddings, facilitating the capture of crucial context and relationships between keywords. This context-aware representation is particularly valuable for the downstream task of Medical Description Generation.
\begin{gather}
\text{align\_score}(ke_s, ke_t) = \bar{ke} = ke_s \odot ke_t \label{eq:align}, \\
ke_{att} = \text{Softmax}(W_{ke}, \bar{ke}) \qquad \forall ke_{att} \in KE_{att}.  \label{eq:att}
\end{gather}
Mathematically, Equation \ref{eq:align} expresses the computation of alignment scores through a dot product operation, while Equation \ref{eq:att} delineates the computation of attention weights using Softmax normalization, where $W_{ke}$ denotes the parameters of the Softmax layer.

\vspace{-3mm}
\subsection{TransFusion Encoder}
\vspace{-2mm}

TransFusion Encoder is the backbone of proposed approach that is designed to receive two types of inputs, lesion contextual attention features and clinical context keyword embeddings from the Visual and Keyword encoders. Main objective of this module is to integrate the clinical context with visual insights from retinal images. It exploits the interactions between visual features and keywords by embedding the keywords using a self-attention mechanism, where attention weights are computed based on the interactions between the Query image $(Q)$ and the Keywords Key $(K)$, Value $(V)$ pairs. This involves a dot product operation between the query and key embeddings, allowing the model to discern the importance of each keyword concerning the retinal image. The attention weights signify the degree of interaction between the image and each keyword, capturing their semantic relationships. Notably, higher attention weights emphasize keywords most relevant to the image, while lower weights assign less importance to less relevant keywords.

\vspace{-5mm}
\begin{equation}
    Q = W_q F_{att} \label{eq:Q}
\end{equation}
\vspace{-5mm}
\begin{equation}
    K = W_k KE_{att} \label{eq:K}
\end{equation}
\vspace{-5mm}
\begin{equation}
    V = W_v KE_{att} \label{eq:V}
\end{equation}
\vspace{-5mm}
\begin{equation}
    \text{Self Attention}(Q, K, V) = Z = \text{Softmax}\left(\frac{QK^{T}}{\sqrt{d_k}}\right)V \label{eq:self-att}
\end{equation}
\vspace{-2.5mm}
\begin{equation}
    Z_{norm} = \text{Layer Normalization}(Q + Z) \label{eq:ln1}
\end{equation}
\vspace{-5mm}
\begin{equation}
    H = W_2 \Gamma(W_1, Z_{norm}) \label{eq:FFNN}
\end{equation}
\vspace{-5mm}
\begin{equation}
    F^{'} = \text{Layer Normalization}(H + Z_{norm}) \label{eq:ln2}
\end{equation}

The mathematical operations underlying the computation of Multi-Head Attention weights $Z$ are detailed in Equations \ref{eq:Q} to \ref{eq:self-att}. Here, $W_q, W_k, W_v$ represent the parameters associated with Attention Features and Context-aware Keyword Embeddings. The dot product operation in Equation \ref{eq:self-att} calculates keyword weights on the image query, with the result scaled by the length of the $K$ vector $(d_k)$. The softmax normalized ratio is then multiplied with $V$ to produce attention probability weights. The subsequent operations involving Residual Addition, Layer Normalization (Equations \ref{eq:ln1} and \ref{eq:ln2}), and Feed Forward Neural Network (FFNN) (Equation \ref{eq:FFNN}) contribute to refining the attention representations, resulting in $F^{'}$.

\vspace{-3mm}
\subsection{Medical Description Decoder}
\vspace{-2mm}

The Medical Description Generation Decoder is a language model based on Transformer Decoder architecture \cite{vaswani2017attention}. It operates as a function of $(F^{'}, E_c, PE_c)$, representing Attention features, Clinical Description Embeddings, and Positional Embeddings (token embeddings + position embeddings), respectively. It comprises of two Masked Multi-Head Attention instances, the first enables self-attention, facilitating the model to focus on different segments of the input sequence while generating the clinical description. The second instance involves cross-attention, allowing the decoder to attend to the encoded retinal image features ($F^{'}$) and comprehend their relevance in generating subsequent words in the clinical description. During self-attention, a Masking (M) operation prevents the model from considering future words, ensuring generation based solely on preceding words, maintaining auto-regressive language generation. Following the attention mechanisms, a feed-forward neural network with two fully connected layers processes the attended and contextualized representations to generate the next word in the clinical description. To stabilize the learning process, residual addition and Layer Normalization are applied post-attention and feed-forward neural network operations. Residual connections enable the model to bypass certain layers, retaining valuable information, while Layer Normalization normalizes outputs, enhancing training stability and overall decoder performance.

\vspace{-5mm}
\begin{equation}
CE = E_c + PE_c \label{eq:CE}
\end{equation}
\vspace{-5mm}
\begin{equation}
Z^{'} = \text{Multi Head Attention}(CE, CE, CE, \text{M}) \label{eq:Z1}
\end{equation}
\vspace{-5mm}
\begin{equation}
Z^{'}_{\text{norm}} = \text{Layer Normalization}(CE + Z^{'}) \label{eq:dln1}
\end{equation}
\vspace{-5mm}
\begin{equation}
Z^{''} = \text{Multi Head Attention}(F^{'}, Z^{'}{\text{norm}}, Z^{'}{\text{norm}}, \text{M}) \label{eq:Z2}
\end{equation}
\vspace{-5mm}
\begin{equation}
Z^{''}{\text{norm}} = \text{Layer Normalization}(Z^{''} + Z^{'}{\text{norm}}) \label{eq:dln2}
\end{equation}
\vspace{-5mm}
\begin{equation}
H^{'} = W_4 \text{ReLU}(W_3, Z^{''}_{\text{norm}}) \label{eq:DFFNN}
\end{equation}
\vspace{-5mm}
\begin{equation}
R^{'}{\text{final}} = \text{Layer Normalization}(H^{'} + Z^{''}{\text{norm}}) \label{eq:dln3}
\end{equation}
\vspace{-5mm}
\begin{equation}
P = \text{Softmax}(R^{'}_{\text{final}}) \qquad P \in (p_0, p_1, p_2, ..., p_m) \label{eq:pred}
\end{equation}

Equations \ref{eq:CE} to \ref{eq:pred} encapsulate the mathematical expressions for generating clinical descriptions based on provided inputs. The resulting probability vector $P$ signifies the probabilities of predicted words in the final clinical description.

The end-to-end training of the Multi-Modal Medical Transformer involves minimizing the cross-entropy loss function $\mathcal{L}(P, C)$, as represented in equation \ref{eq:loss}. This loss function quantifies the dissimilarity between predicted clinical description probabilities $(P)$ and the actual clinical description $(C)$. Summation over all vocabulary words (M), weighted by corresponding actual clinical description values $(c_i)$, is performed. The logarithm of the predicted probability $\log(p_{i})$ ensures more substantial errors are penalized accordingly.
\begin{equation}
\mathcal{L}(P, C) = -\sum_{i=1}^{M} c_{i} \cdot \log(p_{i}) \label{eq:loss}
\end{equation}

This loss function provides a training signal, guiding the model to minimize it, enhancing the capability to generate accurate medical descriptions for given retinal scan images.


\begin{table*}[!ht]
  \centering
  \caption{Results demonstrating the effect of incorporating Keywords as input to the model and integrating Lesion Contextual Gate with Visual features, along with Context-aware Attention to Keyword Embeddings;}
  \label{tab:ablation_results_1}
  \begin{tabular}{cccccccccc}
    \toprule
    \multirow{2}{4em}{Image} & \multirow{2}{4em}{Visual Attention} & \multirow{2}{4em}{Keywords} & \multirow{2}{4em}{Keyword Attention} & \multicolumn{4}{c}{BLEU Scores} & \multicolumn{2}{c}{Other Metrics} \\
    \cmidrule(lr){5-8} \cmidrule(lr){9-10}
    & & & &  BLEU@1 & BLEU@2 & BLEU@3 & BLEU@4 & CIDEr & ROUGE \\
    \midrule
    \checkmark & \xmark & \xmark & \xmark & 0.114 & 0.067 & 0.054 & 0.026 & 0.296 & 0.197 \\
    \checkmark & \checkmark & \xmark & \xmark & 0.184 & 0.110 & 0.068 & 0.032 & 0.316 & 0.230 \\
    \checkmark & \checkmark & \checkmark & \xmark & 0.214 & 0.152 & 0.105 & 0.072 & 0.349 & 0.296 \\
    \checkmark & \checkmark & \checkmark & \checkmark &  0.394 & 0.312 & 0.291 & 0.208 & 0.537 & 0.493 \\
    \bottomrule
  \end{tabular}
  \vspace{-5mm}
\end{table*}

\begin{table*}[!ht]
  \centering
  \caption{Comparative Study of Recent Best Models with Proposed Context Gating Vision Transformer trained on DeepEyeNet Dataset, highlighting the impact of incorporating keyword embeddings and Attention on Visual features;}
  \label{tab:comparision_study}
  \begin{tabular}{lcccccc}
    \toprule
    \multirow{2}{5em}{Model} & \multicolumn{4}{c}{BLEU Scores} & \multicolumn{2}{c}{Other Metrics} \\
    \cmidrule(lr){2-5} \cmidrule(lr){6-7}
    & BLEU@1 & BLEU@2 & BLEU@3 & BLEU@4 & CIDEr & ROUGE \\
    \midrule
    DeepOpth \cite{huang2021deepopht} & 0.184 & 0.114 & 0.068 & 0.032 & 0.361 & 0.232 \\
    Deep Context Encoding Network \cite{huang2021deep} & 0.219 & 0.134 & 0.074 & 0.035 & 0.398 & 0.252 \\
    Contextualized Keywords \cite{huang2021contextualized} & 0.203 & 0.142 & 0.100 & 0.073 & 0.389 & 0.211 \\
    Non-local Attention \cite{huang2022non} & 0.230 & 0.150 & 0.094 & 0.053 & 0.370 & 0.291 \\
    \textbf{M3 Transformer (Ours)} & \textbf{0.394} & \textbf{0.312} & \textbf{0.291} & \textbf{0.208} & \textbf{0.537} & \textbf{0.493} \\
    \bottomrule
  \end{tabular}
  \vspace{-5mm}
\end{table*}

\vspace{-3mm}
\section{Experiments \& Results}
\label{sec:experiments}
\vspace{-2mm}

\subsection{Dataset} \label{ssec:dataset}
\vspace{-2mm}

The DeepEyeNet dataset, introduced in \cite{huang2021deepopht}, is a large-scale collection of retinal images comprising 15,709 scans obtained from diverse imaging modalities such as Fluorescein Angiography (1,811 images), Fundus Photography (13,898 images), and Optical Coherence Tomography (OCT). Each image is meticulously annotated by expert ophthalmologists, providing multiple diagnostic keywords that capture medical observations and clinical descriptions related to the retinal findings. Keyword and clinical description lengths vary, with the longest keywords exceeding 15 words and the longest clinical descriptions surpassing 50 words. On average, both keywords and clinical descriptions have a word length ranging from 5 to 10 words. The dataset covers 265 distinct retinal diseases and symptoms, encompassing a broad spectrum of both common and less common conditions. This dataset is divided into standard splits: 60\% for training (9,425), 20\% for validation (3,142), and 20\% for testing (3,142).

\vspace{-3mm}
\subsection{Pre-processing}
\vspace{-2mm}
Effective pre-processing is a foundation for optimizing the quality and interpretability of retinal images across diverse modalities such as color fundus, fundus auto-fluorescence, and OCT. In our approach, we implement crucial steps on acquired retinal scan images to mitigate variations in size, dimensions, and color arising from different devices and laboratory conditions. This includes resizing operations to standardize images to a uniform scale, as well as converting images into a three-channel format (Red, Green, Blue) to enhance processing flexibility and compatibility with pre-trained neural networks. Simultaneously, for clinical descriptions and diagnostic keywords, essential text processing steps are applied, encompassing removal of non-alphabetic characters, conversion to lowercase for uniformity, and handling rare words by replacing them with a UNK token. The list of provided keywords is transformed into a sequence by replacing comma separators with [SEP] during preprocessing.

\vspace{-3mm}
\subsection{Experimental Settings}
\vspace{-2mm}
In our study, experiments were conducted on the DeepEyeNet dataset, adopting established settings from prior works \cite{huang2021deepopht, huang2021deep, huang2021contextualized, huang2022non}. To ensure comparability, consistent experimental conditions were maintained, and evaluation metrics (BLEU, CIDEr, ROUGE) aligned with those in referenced studies. The experiments were performed on an Nvidia P100 GPU (16GB Memory, 1.32GHz clock, 9.3 TFLOPS). Optimizing our model's performance involved carefully chosen hyperparameters, with a learning rate of 0.004, Adam optimizer, batch size of 64, and dropout rate of 0.2. The model architecture comprises a two-layer fully connected feed-forward neural network (FFCN) with hidden layer sizes of 512 and 256, demonstrating tailored efficiency. A vocabulary size of 5000, 300-dimensional word embeddings, and set sequence lengths (5 to 50 words) contribute to the model's effectiveness. Categorical cross-entropy as the loss function solidifies our hyperparameter choices for successful model training.

\begin{table*}[!t]
  \centering
  \small
  \caption{Comparison of Actual, Non-Local Attention, and M3T medical descriptions}
  \label{tab:clinical_desc_results}
  \begin{tabular}{cccccc}
    \toprule
    \textbf{Retinal Image} & \textbf{Keywords} & \textbf{Groundtruth} & \textbf{Non-Local Attention \cite{huang2022non}} & \textbf{M3 Transformer} \\
    \midrule
    \multirow{8}{*}{\includegraphics[width=2.5cm, height=2.5cm]{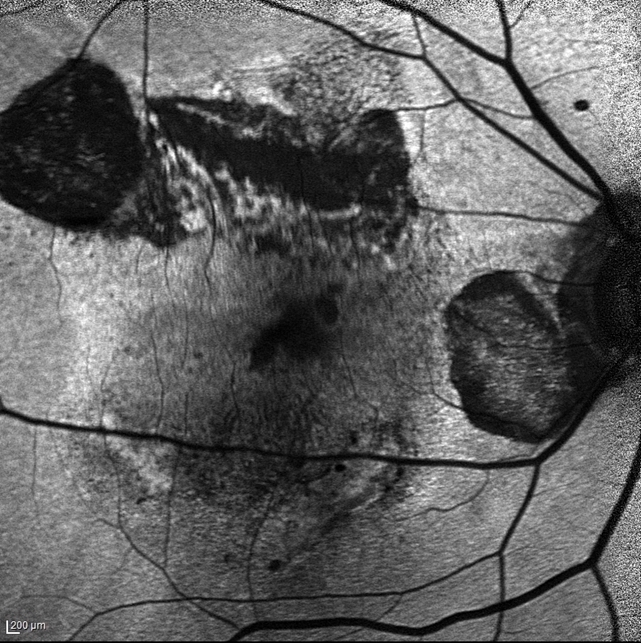}} & \multirow{8}{7em}{autofluorescence imaging[sep]age-related macular degeneration (amd)} & \multirow{9}{12em}{an autofluorescence image of a 78 year old man with an age related macular degeneration on his both eyes} & \multirow{8}{12em}{Autofluorescence to image of the right eye of a [age] year old [gender] with acute decrease in vision mainly right eye is with pigment clumping and optic nerve drusen in the right eye} & \multirow{8}{12em}{autofluorescence image of a 76 year old woman with age related macular degeneration on her right eye and the left eye} \\
     &  &  &  & \\
     &  &  &  & \\
     &  &  &  & \\
     &  &  &  & \\
     &  &  &  & \\
     &  &  &  & \\
     &  &  &  & \\
    \midrule
    \multirow{9}{*}{\includegraphics[width=2.5cm, height=2.5cm]{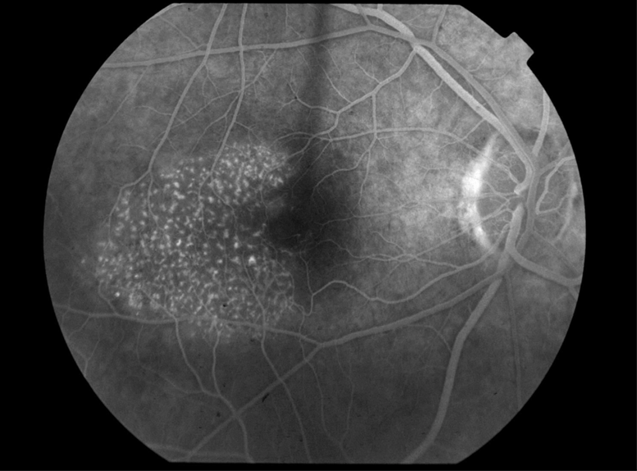}} & \multirow{9}{7em}{idiopathic macular telangiectasia[sep]parafoveal telangiectasia[sep]juxtafoveal telangiectasis} & \multirow{9}{12em}{the fellow eye was unremarkable on this red free image} & \multirow{9}{12em}{The telangiectasis occurs unilaterally in the macula be of choroidal folds peripheral nevus shows cnvm in this eyes shows resolved with myopia over the macula and lung remained and remained over the exam was remained cnvm. Resolved and pdt. } & \multirow{9}{12em}{the telangiectasis occurs unilaterally in the temporal half of the macula in an area of 1–2 disc diameters the anomalies are noted on this red free image in this early frame of the angiogram} \\
     &  &  &  & \\
     &  &  &  & \\
     &  &  &  & \\
     &  &  &  & \\
     &  &  &  & \\
     &  &  &  & \\
     &  &  &  & \\
     &  &  &  & \\
    \bottomrule
  \end{tabular}
  \vspace{-5mm}
\end{table*}

\vspace{-3mm}
\subsection{Quantitative Evaluation} \label{ssec:quant_evaln}
\vspace{-2mm}
In the ablation study in Table \ref{tab:ablation_results_1}, we meticulously dissect the components of our M3 Transformer to evaluate their individual and cumulative impacts on medical description generation for retinal images. Notably, the progressive inclusion of visual attention, keywords, and attention mechanisms on both visual and keyword features leads to an improvement in BLEU scores. The results signify the benefit of visual attention in focusing on relevant regions within retinal images, while the incorporation of diagnostic keywords significantly enhances the model's ability to integrate clinical context. The culmination of these components in the full model demonstrates a substantial leap in BLEU scores, reinforcing the synergistic effect of jointly considering visual and clinical information for robust medical description generation.

In the comparative study illustrated in Table \ref{tab:comparision_study}, our proposed M3 Transformer Network is pitted against recent leading models in the field of retinal image clinical description generation. Against models like DeepOpth, Deep Context Encoding Network, Contextualized Keywords, and Non-local Attention, our M3 Transformer exhibits superior performance across all BLEU scores, showcasing its capacity to generate more accurate and coherent clinical descriptions. These results highlight the distinctive strength of our approach in seamlessly integrating multi-modal information from retinal images and clinical keywords, making significant strides in addressing the challenges of generating clinically relevant descriptions for diverse retinal conditions.

\vspace{-3mm}
\subsection{Qualitative Evaluation} \label{ssec:qual_evaln}
\vspace{-2mm}

\begin{figure}[tb!]
    \centering
    \centerline{\includegraphics[width=.95\columnwidth]{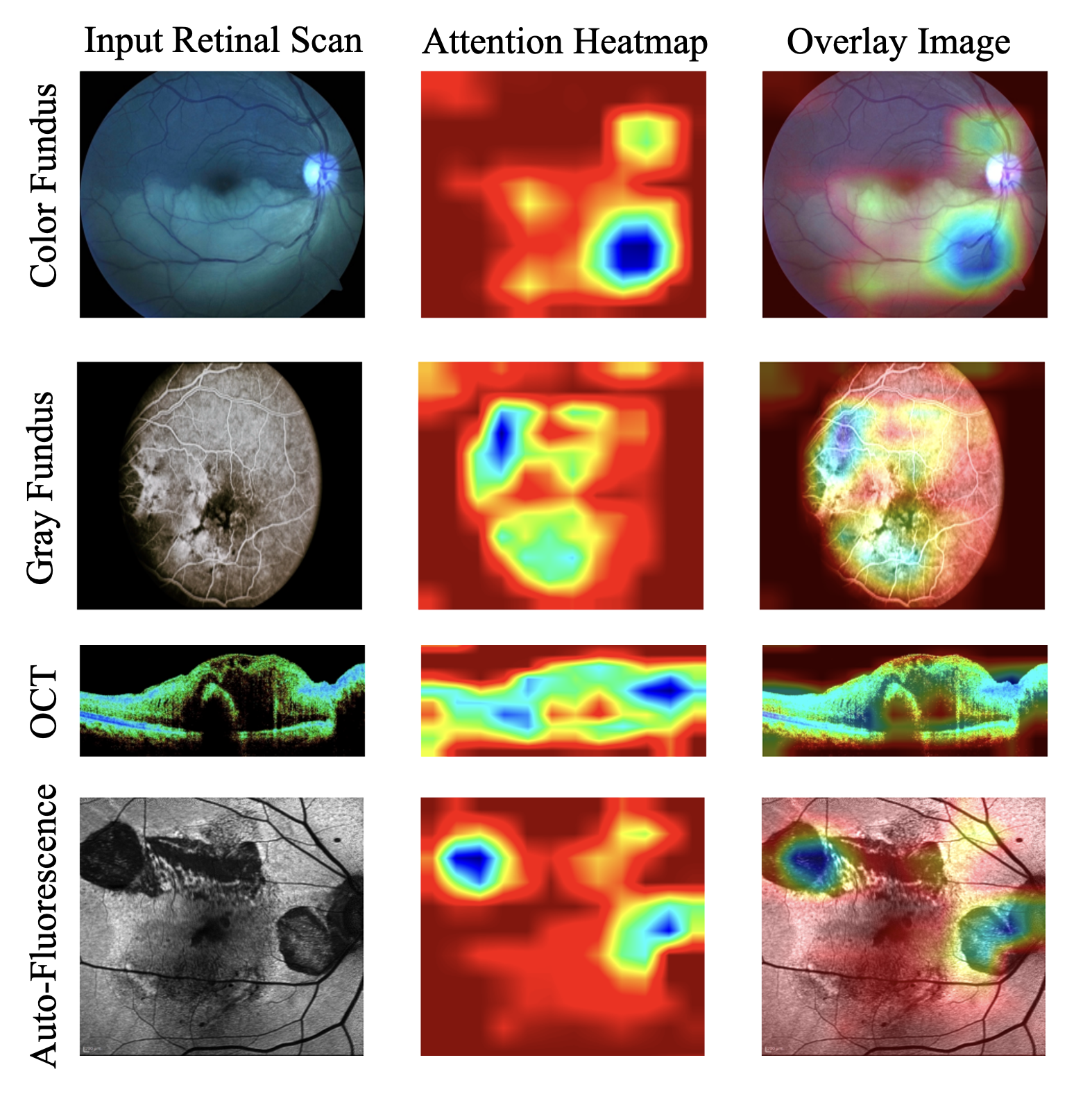}}
    \vspace{-5mm}
    \caption{Visualization of Lesion Contextual Gate Attention heatmaps and corresponding overlays, highlighting the visual insights from color fundus, gray fundus, OCT, and Auto-Fluorescent retinal scan images}
    \vspace{-4mm}
    \label{fig:att-maps}
\end{figure}

In Figure \ref{fig:att-maps}, we provide a visual representation of the Lesion Contextual Gate Attention heatmaps and corresponding overlays, showcasing the explainability of our proposed approach. The heatmaps highlight the visual insights extracted from diverse retinal imaging modalities, including color fundus, gray fundus, optical coherence tomography (OCT), and autofluorescent retinal scan images. The Lesion Contextual Gate effectively focuses on relevant regions within the retinal images, emphasizing areas crucial for clinical understanding. This visualization underscores the model's capability to integrate information from various modalities, providing ophthalmologists with interpretable indications of the visual cues influencing the generated clinical descriptions.

Table \ref{tab:clinical_desc_results} offers a qualitative comparison of actual, Non-Local Attention \cite{huang2022non}, and predicted medical descriptions using retinal images and keywords as input. Our M3 Transformer outperforms the Non-Local Attention model by generating more accurate and coherent clinical descriptions. For instance, in the first case, the Non-Local Attention model produces a description with generalities about age and gender, while our M3 Transformer provides specific details about macular hole conditions in both eyes. Similarly, in the second case, our model accurately identifies and describes the presence of idiopathic macular telangiectasia, parafoveal telangiectasis, and juxtafoveal telangiectasis, offering detailed insights into the retinal anomalies. This comparison highlights the superior performance of our proposed approach in capturing nuanced clinical information and generating clinically relevant descriptions with greater accuracy and specificity.

\vspace{-3mm}
\section{Conclusion} \label{sec:conclusion}
\vspace{-2mm}

This research introduces the Multi-Modal Medical Transformer (M3T) for generating precise medical descriptions from retinal images, addressing the rising global prevalence of visual impairments. The proposed model integrates diagnostic keywords and visual features, demonstrating superior performance compared to existing models in the field. Through an ablation study, we highlight the incremental contributions of visual attention, keywords, and attention mechanisms, emphasizing their collective impact on accurate medical description generation. Quantitative evaluations confirm the effectiveness of M3T, surpassing state-of-the-art models in terms of all metrics reported. Additionally, qualitative assessments demonstrate the model's explainability through attention heatmaps. In future, M3T model can be further extended to diverse medical imaging domains and enhancing its interpretability for broader clinical applicability.

\bibliographystyle{IEEEbib}
\bibliography{refs}

\end{document}